\documentclass[journal]{IEEEtran}
\usepackage{cite}
\usepackage{verbatim}
\usepackage{graphicx}
\usepackage[table]{xcolor}
\usepackage{textcomp}
\usepackage[margin={0.7in}]{geometry}
\usepackage{amsmath,amssymb,amsfonts,bm}
\usepackage{booktabs}
\usepackage{epsf}
\usepackage{tikz}
\usepackage{psfrag}
\usepackage{pstool}
\usepackage{epstopdf}
\usepackage{stfloats}

\usepackage{url}
\usepackage[normalem]{ulem}
\usepackage[dvips]{epsfig}
\usepackage{amsthm}
\usepackage{color}
\usepackage[usenames,dvipsnames]{pstricks}
\usepackage{lettrine}
\usepackage[dvips]{epsfig}
\usepackage{pst-grad} 
\usepackage{pst-plot} 
\usepackage{epsfig}
\usepackage{enumerate}
\usepackage{amsbsy}
\usepackage{amssymb}
\usepackage{amsthm}
\usepackage{amscd}
\usepackage{float}
\usepackage[caption = false]{subfig}
\usepackage{color}
\usepackage[numbers,sort&compress,square]{natbib}

\usepackage{listings}
\usepackage{stackrel}
\usepackage{authblk}
\usepackage{dsfont}
\usepackage{algorithm}
\usepackage[noend]{algpseudocode}
\makeatletter
\newcommand{\removelatexerror}{\let\@latex@error\@gobble}
\makeatother

\usepackage{array}
\newcolumntype{L}[1]{>{\raggedright\let\newline\\\arraybackslash\hspace{0pt}}m{#1}}
\newcolumntype{C}[1]{>{\centering\let\newline\\\arraybackslash\hspace{0pt}}m{#1}}
\newcolumntype{R}[1]{>{\raggedleft\let\newline\\\arraybackslash\hspace{0pt}}m{#1}}
\usepackage[subtle]{savetrees}

\begin{document}

\ifCLASSINFOpdf

\fi

\def\BibTeX{{\rm B\kern-.05em{\sc i\kern-.025em b}\kern-.08em
    T\kern-.1667em\lower.7ex\hbox{E}\kern-.125emX}}
 \setlength{\columnsep}{0.21in}

\title{{Federated Learning in NTNs: Design, Architecture and Challenges}}

\author{
Amin Farajzadeh, Animesh Yadav, and Halim Yanikomeroglu
\thanks{A. Farajzadeh and H. Yanikomeroglu are with the Non-Terrestrial Networks (NTN) Lab, Department of Systems and Computer Engineering, Carleton University, Ottawa, ON K1S 5B6, Canada. A. Yadav is with the School of EECS, Ohio University, Athens, OH, 45701 USA.}}

\makeatletter
\patchcmd{\@maketitle}
  {\addvspace{0.5\baselineskip}\egroup}
  {\addvspace{-1.4\baselineskip}\egroup}
  {}
  {}
\makeatother

\maketitle

\IEEEpeerreviewmaketitle

\begin{abstract}
Non-terrestrial networks (NTNs) are emerging as a core component of future 6G communication systems, providing global connectivity and supporting data-intensive applications. In this paper, we propose a distributed hierarchical federated learning (HFL) framework within the NTN architecture, leveraging a high altitude platform station (HAPS) constellation as intermediate distributed FL servers. Our framework integrates both low-Earth orbit (LEO) satellites and ground clients in the FL training process while utilizing geostationary orbit (GEO) and medium-Earth orbit (MEO) satellites as relays to exchange FL global models across other HAPS constellations worldwide, enabling seamless, global-scale learning. The proposed framework offers several key benefits: (i) enhanced privacy through the decentralization of the FL mechanism by leveraging the HAPS constellation, (ii) improved model accuracy and reduced training loss while balancing latency, (iii) increased scalability of FL systems through ubiquitous connectivity by utilizing MEO and GEO satellites, and (iv) the ability to use FL data, such as resource utilization metrics, to further optimize the NTN architecture from a network management perspective. A numerical study demonstrates the proposed framework’s effectiveness, with improved model accuracy, reduced training loss, and efficient latency management. The article also includes a brief review of FL in NTNs and highlights key challenges and future research directions.
      
\end{abstract}
\vspace{-3mm}
\section{What Drives NTN for 6G and beyond?}

The sixth generation (6G) wireless networks aim to provide global connectivity, immersive, hyper-reliable, low-latency communications, and support a range of new applications across industries. Traditional terrestrial networks alone cannot meet these ambitious goals. As a result, researchers are focusing on developing non-terrestrial networks (NTNs) to create a seamless, integrated global network that serves both remote and urban areas. NTNs represent a key advancement in achieving these objectives and enabling novel applications. NTNs integrate three main network tiers—terrestrial, aerial, and satellite—with the aerial and satellite tiers executing the core functions of resource management, coordination, and robust connectivity, while the terrestrial tier supports edge operations. The terrestrial tier comprises ground-based entities such as base stations and users. The aerial tier consists of flying vehicles, including unmanned aerial vehicles (UAVs) and high-altitude platform stations (HAPS) operating at altitudes up to 22 kilometres. The satellite tier encompasses satellites in geostationary (GEO), medium-Earth (MEO), and low-Earth (LEO) orbits.

NTNs are considered not only to serve rural and remote areas, such as islands, seas, and deserts, but also to offer services in urban environments, including immersive communications, massive connectivity, edge computing, smart cities, and intelligent transportation systems. Leveraging the large footprint and line-of-sight (LoS) links, satellites and HAPS provide broadband to remote users and high-capacity direct links to urban users, as well as relay links between satellites and ground stations \cite{kurt2021IEEEsnt}. Due to heterogeneous network entities, NTNs offer more resources, multiple connectivity options, routing paths, and large-scale, stable connectivity for users in both rural and urban areas \cite{FL-SAGINE-IEEE-NetMag-2023}.
\vspace{-3.5mm}
\section{Compatibility of FL and NTN}
Federated learning (FL) is a collaborative learning approach in which global training is carried out on distributed private raw data without sharing it. FL was largely developed with terrestrial network designs in mind. However, the FL framework is perfectly suited to NTN's complicated and multi-tier structure in two ways: (i) FL for NTN and (ii) NTN for FL. 

\textbf{FL for NTN:} FL can be used to empower NTNs by leveraging learning‐based optimization methods to tackle issues such as user scheduling and offloading, resource and interference management, channel estimation, and optimization of aerial nodes placement and trajectory. By doing so, FL enables NTNs to dynamically adapt to heterogeneous network conditions and diverse client capabilities, ultimately improving performance even in remote or challenging environments. Obtaining the optimal decision with low latency and dependability in multi-tier heterogeneous networks is challenging due to their diverse functionalities and connectivity possibilities \cite{FL-SAGINE-IEEE-NetMag-2023}. Machine learning (ML)-based solutions can help us due to their powerful ability to approximate optimization problems and network functions faster and accurately \cite{Dahrouji-ML-for-Optimization-IEEE-Access-2021}. Traditional centralized ML methods require extensive data sharing, which leads to network congestion, delays, and potential data leakage, making FL a more efficient and secure alternative for optimizing NTN performance.

\textbf{NTN for FL:} The essence of NTN for FL lies in providing a robust, multi-tier infrastructure that enhances FL performance. The multi-tier NTN architecture—comprising terrestrial, aerial, and satellite tiers—facilitates the aggregation of model updates from a vast and diverse pool of clients. This extensive connectivity not only improves model accuracy and scalability but also bolsters privacy by ensuring that only aggregated updates are shared. In essence, NTNs extend the reach of FL to regions beyond conventional networks, enabling more reliable and resilient distributed learning. Satellites and ground devices collect vast amounts of high-resolution data. Typically, satellite data is downloaded to a parameter server (PS) on Earth, but limited bandwidth makes this process a bottleneck. The FL framework eliminates the need for data downloads, though challenges like intermittent connectivity and varying satellite orbits can cause model staleness and imbalanced participation. In a multi-tier NTN architecture, intermediate entities like HAPS can act as the PS (or FL server), addressing these issues \cite{FedHAP-2022}. Additionally, FL and NTN enhance IoT networks by involving multiple devices across different locations, improving model accuracy and convergence \cite{FLSTRA-2024}. Thus, NTN for FL holds great promise for advancing satellite and IoT networks.

Despite this potential, most studies focus on centralized FL between two or within the same tier of NTN. To fully leverage NTN, all three tiers must be considered. Furthermore, improving model aggregation and accuracy while minimizing latency requires a decentralized, scalable, and resilient learning environment. To this end, we introduce a fully distributed hierarchical FL (HFL) framework, utilizing HAPS constellation nodes as distributed FL servers. In the following, we will present the proposed NTN architecture, review state-of-the-art FL literature in NTN, and discuss the performance evaluation of the distributed HFL framework. Lastly, we address deployment challenges and future directions.  
\begin{figure*} [!t]
    \centering
    \includegraphics[width=162mm]{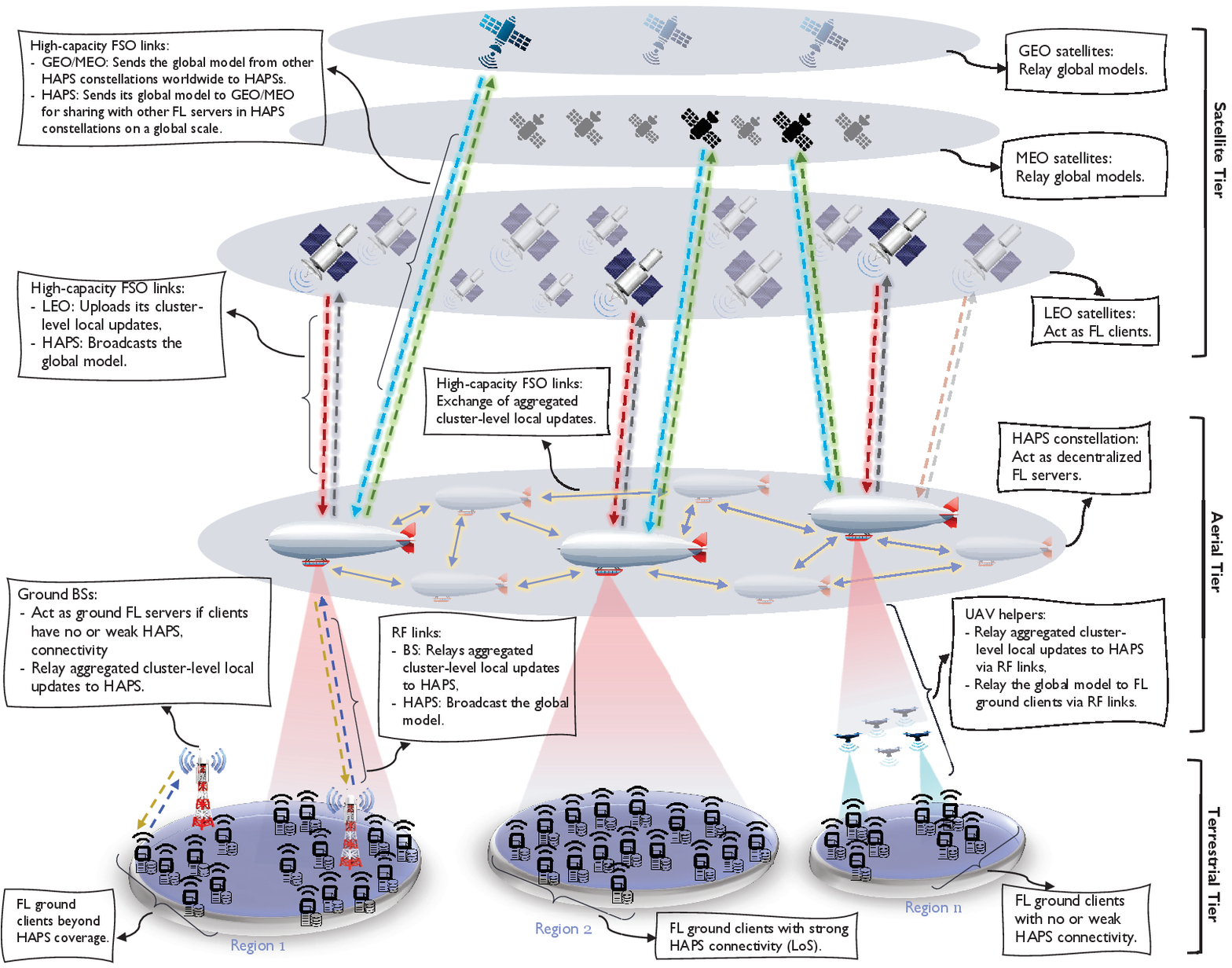}
    \caption{The proposed NTN architecture for distributed HFL.}
    \label{fig:sys_model}
\end{figure*}
\vspace{-2mm}
\section{NTN Architecture: An Overview}
The proposed NTN architecture comprises three primary tiers: satellite, aerial, and terrestrial, as depicted in Fig.~\ref{fig:sys_model}. Each tier plays a crucial role in enabling distributed, ubiquitous, and high-performance FL mechanisms across NTNs.
\vspace{-4mm}
\subsection{Satellite Tier}
The satellite tier consists of GEO, MEO, and LEO satellites, providing overarching connectivity within the network. Despite their significant coverage area, satellites are less effective for low-latency, low-power communication due to their high altitude and distance from ground devices. Their primary functions are twofold:
\begin{enumerate}
    \item \textbf{Satellite Clients:} LEO satellites perform low-power computing tasks with a limited number of iterations, making them potential FL clients. The LEO mega-constellation enables the participation of numerous LEO satellites in the FL system, facilitating fast and reliable communication with lower network tiers to share or exchange model updates, which contribute to generating new global models or parameters.
    \item \textbf{Satellite Relays:} GEO and MEO satellites, with their extensive coverage but limited computational resources, act primarily as relays. They receive global models from one or more HAPS nodes within a constellation and relay them to geographically distant HAPS nodes in other constellations, enabling a ubiquitous FL system on a global scale.
\end{enumerate}
\vspace{-4mm}
\subsection{Aerial tier}
The aerial tier consists mainly of HAPS nodes and helper UAVs. HAPS nodes serve as primary intermediate aggregators, offering an optimal balance between coverage and latency. UAVs act as temporary helpers to relay and forward updates between the terrestrial tier and HAPS nodes, ensuring flexibility and robustness in data transmission. HAPS nodes mitigate the high latency and high power consumption issues associated with direct satellite communication, serving as key nodes for distributed aggregation.
\begin{enumerate}
    \item \textbf{Distributed FL Servers:} 
HAPS nodes, strategically located in the stratosphere, offer a unique advantage for distributed FL systems by bridging the satellite and terrestrial tiers with stable, low-latency communication. Their positioning enables efficient aggregation and model updates without high delays associated with satellite-only systems. The flexibility of designing a HAPS constellation allows a group of multiple HAPS to act as decentralized FL servers, leveraging substantial computational resources to manage complex FL tasks, making fully distributed learning both feasible and efficient.
    \item \textbf{Aerial Relays:} UAVs provide additional flexibility and can act as temporary relays, particularly useful in dynamically changing network conditions. They support HAPS nodes by relaying data from ground devices and BSs, especially in areas with high mobility or during peak traffic conditions. UAVs can dynamically adjust their routes based on real-time network demands, ensuring efficient data collection and transmission.
\end{enumerate}
\vspace{-5mm}
\subsection{Terrestrial tier}
The terrestrial tier consists of ground-edge devices and BSs.
\begin{enumerate}
    \item \textbf{Ground Clients:} Ground-edge devices, such as cell phones, sensors, cameras, and smart appliances, generate large volumes of data. These devices are typically equipped with communication and computational hardware capable of running learning algorithms, such as convolutional neural networks (CNNs), and transmitting and receiving communication signals.
    \item \textbf{Ground FL Servers and Ground Relays:} When direct connectivity between ground clients and a HAPS node is unavailable, BSs act as aggregation points. They collect local updates from ground clients and forward them to the aerial tier for further processing. If BSs are outside the HAPS coverage, they function as FL servers, handling model generation and broadcasting, similar to conventional FL systems.
\end{enumerate}
\vspace{-3.5mm}
 \subsection{Why is HAPS Constellation Ideal for Distributed FL Systems?}
A HAPS constellation offers a powerful solution for distributed FL due to its optimal balance between coverage, latency, and computational capability. Positioned in the stratosphere, HAPS nodes provide stable, low-latency communication between satellite and terrestrial tiers, which is crucial for seamless model updates. Unlike satellites, HAPS nodes can maintain constant connectivity and handle complex aggregation tasks, and together they form decentralized learning servers. Equipped with high-capacity links, such as free-space optical (FSO) communication, supported by strong LoS connections, HAPS nodes ensure fast, and reliable data transmission. Their flexible constellation design makes them ideal for scalable, fully distributed FL across diverse network environments, as discussed in Section V.

\begin{table*}[!t]
\centering
\caption{Summary of FL Features and Main Goals for Different Tiers in NTN Architecture}
\vspace{-2mm}
{\rowcolors{2}{white}{black!10}
\begin{tabular}{|l|p{7.5cm}|p{8cm}|}
\hline
\textbf{Tier} & \textbf{FL Features} & \textbf{Main Functionalities (or Goals)} \\ \hline \vspace{-2mm}
Terrestrial & \vspace{-2mm}
\begin{itemize}
    \item Ground clients: Ground-based devices (e.g., sensors, IoT devices).
    \item Ground FL servers or relays: Ground BSs.
    \item Communication links: RF for all ground-to-ground and ground-to-HAPS/UAV connections.
\end{itemize}
& \vspace{-2mm}
\begin{itemize}
    \item Local training and data processing at each client in the cluster.
    \item Relaying cluster-level updates to HAPS via ground BSs.
    \item Aggregation of updates at BSs where HAPS connectivity is weak.
\end{itemize}
 \\ \hline \vspace{-2.5mm}
Aerial & \vspace{-2mm}
\begin{itemize}
    \item Distributed FL servers: HAPS Constellation nodes.
    \item Aerial relays: UAV helper nodes.
    \item Communication links: i) High-capacity FSO for HAPS-to-HAPS connections, ii) RF for HAPS-to-UAV and HAPS/UAV-to-ground communications.
\end{itemize} 
& \vspace{-2.5mm}
\begin{itemize}
    \item Intermediate aggregation of cluster-level updates.
    \item Relaying updates to HAPS via UAV helpers.
    \item P2P exchange of updates among neighbouring HAPS.
    \item Generation and broadcasting of the new global model.
    \item Recording and leveraging FL data for NTN management.
\end{itemize}
 \\ \hline
Satellite & \vspace{-2mm}
\begin{itemize}
    \item Satellite clients: LEO satellites.
    \item Satellite relays: GEO and MEO satellites.
    \item Communication links: High-capacity FSO for both LEO-to-HAPS and GEO/MEO-to-HAPS connections.
\end{itemize}
& \vspace{-2.5mm}
\begin{itemize}
    \item Relay FL models globally to HAPS nodes in the aerial tier.
    \item Integrate new FL clients, such as GEO satellites, into the network.
    \item Ensure ubiquitous connectivity across vast geographical areas.
    \item Achieve global model consistency through long-distance aggregation. 
\end{itemize}
\\ \hline \end{tabular} \vspace{-3mm}
}
\label{tab:summary}
\end{table*}
\vspace{-3mm}
\section{State-of-the-Art FL in NTN}
This section provides a brief literature review of how FL is applied in various NTN configurations followed by the proposed novel framework, which extends FL across all three tiers.
\vspace{-3mm}
\subsection{Satellite-Satellite Networks}
In this network setting, FL is performed within the satellite tier, specifically in a satellite-satellite network \cite{shi2024satellite, Yang2024DFedSatCA, lin2024fedsnnovelfederatedlearning, FedLEO_2024}.
In \cite{shi2024satellite}, the FedMega algorithm, based on federated edge learning, performs FL between LEO mega-constellation satellites using stable, high-data-rate intra-satellite links (ISL). \cite{Yang2024DFedSatCA} introduces DFedSat, a decentralized FL framework for LEO satellites, using orbit reduction and a flexible gossip scheme for partial model aggregation and addressing unreliable inter-orbit links with a self-compensation mechanism.
\cite{lin2024fedsnnovelfederatedlearning} proposes FedSN, an FL framework considering heterogeneous computing, memory, uplink rates, and model staleness in LEO satellites. It allows heterogeneous local model training and introduces a pseudo-synchronous aggregation strategy to mitigate staleness by dynamically scheduling aggregation.
In \cite{FedLEO_2024}, the FedLEO framework enables decentralized FL in LEO satellites. Each satellite trains its local model using the SGD method for several epochs, then performs decentralized aggregation by communicating with adjacent satellites until a global consensus model is achieved.
\vspace{-2mm}
\subsection{Satellite-Terrestrial Networks}
In this network setting, FL operates between satellite and terrestrial tiers to address bandwidth limitations and intermittent connectivity. The FedLEO algorithm in \cite{Elmahallawy2023OptimizingFL} has a visible LEO satellite that shares the global model with neighbouring satellites via ISL. After local training, the models are sent to a sink satellite, which generates a partial global model and downloads it to the FL server when visible. \cite{So2022FedSpaceAE} introduces FedSpace, which schedules global aggregation, based on deterministic satellite connectivity, using Earth's rotation and satellite orbits. \cite{Han2023CooperativeFL} presents a cooperative FL framework where satellites act as edge computing units, aggregators, and relays, coordinating with ground clusters for local and global model aggregation.
\vspace{-2mm}
\subsection{Aerial-Satellite Networks}
FL is performed between aerial and satellite tiers in this network setting. The algorithms proposed in \cite{FedHAP-2022, elmahallawy2022asyncfleoasynchronousfederatedlearning, FedHAP-NOMA_JSAC_2024} tackle the issue of prolonged training time due to challenging channel between satellite and the FL server on the ground. In \cite{FedHAP-2022}, the FedHAP framework that uses HAPS instead of the ground station as the FL server for synchronous FL is proposed to orchestrate a fast and efficient model training process for LEO constellations. However, to overcome the problem of model staleness caused by straggler satellites in FedHAP, AsyncFLEO is proposed in \cite{elmahallawy2022asyncfleoasynchronousfederatedlearning}. Essentially, instead of waiting to collect model parameters from the satellites, the AsyncFLEO framework first groups the satellites based on their data distribution in the first epoch and then, from the group, selects a subset of satellites based on their model freshness for model aggregation in the subsequent epoch.
\vspace{-3.7mm}
\subsection{Aerial-Terrestrial Networks}
In this network setting, FL is performed between HAPS and ground tier \cite{FLSTRA-2024, HAP-DDPG-Hindawi-2023}. In \cite{FLSTRA-2024}, the FLSTRA framework is introduced to enhance convergence rates and reduce communication delays caused by limited user participation and multi-hop communications. The large footprint of HAPS on the ground is leveraged to accommodate many ground users. A joint user selection and resource allocation strategy is developed to minimize the FL delay. In \cite{HAP-DDPG-Hindawi-2023}, the problem of finding optimal aerial base station (AeBS) 3D placement is studied to maximize the link capacity between the ground user and AeBS. To achieve this objective, the HAPS-assisted AeBS deployment architecture is developed via a federated deep reinforcement learning (DRL) framework. Each AeBS trains its local DRL model in this architecture and relays it to the HPAS, which acts as the FL server for global DRL model aggregation.   
\vspace{-3.7mm}
\subsection{Satellite-Aerial-Terrestrial Networks}
In this network setting, FL is performed across satellite, aerial, and terrestrial tiers. In \cite{FL-SAGINE-IEEE-NetMag-2023}, the potential of FL to enhance the control and performance of complex and dynamic NTNs is explored. The role of FL in applications such as resource allocation, dynamic node scheduling, and network traffic management is also discussed.

The integration of FL across all three tiers of NTNs is still infancy. TABLE I summarizes the FL features and main goals of the terrestrial, aerial, and satellite tiers in NTN architecture, highlighting their collective contribution to performance and scalability. Given the advantages and strategic positioning of HAPS within NTNs, we propose a distributed hierarchical learning framework, referred to as distributed hierarchical federated learning (HFL), to facilitate this integration. Details of the framework are provided in the following section.

\vspace{-3mm}
\section{Distributed Hierarchical FL}
The proposed HFL framework leverages the hierarchical structure of the NTN to enable scalable and efficient model training and aggregation. Unlike conventional terrestrial-based HFL frameworks, the proposed NTN-based distributed HFL framework leverages the strategically positioned HAPS constellation to decentralize the aggregation and model generation processes. This approach facilitates the participation of a large number of clients, including ground-edge devices and LEO satellites, from various regions around the globe. In this framework, local models are first aggregated at the cluster level, where each cluster comprises the FL clients (ground users or LEO satellites) within a single HAPS node's coverage. These cluster-level updates are then exchanged among neighbouring HAPS nodes, which generate a global model as a function (e.g., arithmetic, weighted, or logarithmic) of all aggregated updates. Hence, decentralized aggregation via the HAPS constellation obscures individual contributions, mitigating reverse-engineering risks and providing an extra layer of privacy beyond conventional FL practices.

\begin{figure*}[!t]  
    \centering 
    \hspace*{-3mm}
   \includegraphics[width=186mm]{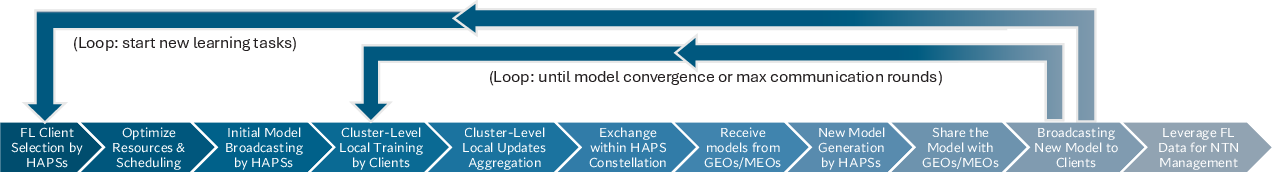} \vspace*{-5mm}
    \caption{The proposed distributed HFL framework.
    } \vspace{-2mm}
    \label{fig2}
\end{figure*}

Besides, the proposed framework offers several other key benefits: (i) improved model accuracy and reduced training loss while balancing latency, (ii) increased scalability to a global scale utilizing MEO and GEO satellites, and (iii) the ability to use FL data, such as resource utilization metrics, after the learning task is complete to optimize the NTN architecture further, enabling ``FL for NTN''. In the following subsections, we detail the key steps involved in the proposed HFL framework. The flowchart summary of the proposed HFL framework is depicted in Fig.~\ref{fig2}.

\vspace{-2mm}
\vspace{-2.5mm}
\subsection{FL Client Selection}
Each HAPS node in the aerial tier forms a cluster of clients by selecting a set of potential LEO satellites and ground-edge devices in the satellite and terrestrial tiers, respectively, to participate in the cluster-level FL training process. A ground user is a potential client if it has a sufficient signal-to-noise ratio, regardless of LoS or NLoS. If connectivity to HAPS is weak, updates are aggregated by a nearby BS or relayed via helper UAVs. In contrast, only LEO satellites visible to HAPS are selected as clients. FSO links connect HAPS to LEO, while RF links connect HAPS to the ground. For FL training, the final list of clients is chosen based on computing power, communication capabilities, and resource availability to ensure efficiency. 

A key challenge in distributed FL over NTN architectures is the non-IID nature of data among clients, caused by differences in environments, sensor types, and user behaviours, affecting model convergence and accuracy. To address this challenge, clients are strategically selected based on similar characteristics (e.g., human-type or machine-type) and traffic histories~\cite{Elmahallawy2023OptimizingFL}.
\vspace{-3mm}
\subsection{Optimizing Resource Allocation and Scheduling}
Limited energy of clients, including ground and LEO clients, and scarce network resources, such as bandwidth, can significantly impact the convergence of the learning process. Effective resource allocation and client scheduling are crucial for completing FL tasks efficiently. Before training starts, key resources like bandwidth, CPU frequency, and transmit power are optimized to balance energy use, learning accuracy, and timing. Proper management allows devices to engage in learning without draining their energy. Dynamic scheduling algorithms then prioritize transmissions based on resource availability and channel conditions, ensuring only high-quality model updates are sent to intermediate nodes, which enhances learning and overall network performance.
\vspace{-3.5mm}
\subsection{Model Broadcasting and Cluster-Level Local Training}
Each HAPS node broadcasts the global model to the clients within its cluster, where clients utilize their private datasets for cluster-level local training. These selected clients independently update their models, either synchronously or asynchronously, using task-specific algorithms (e.g., CNNs). Once training is complete or a predefined number of epochs is reached, these clients upload their refined model updates to the associated HAPS for aggregation. This sequential process produces cluster-level local updates that are subsequently used to update the global model.

\vspace{-3.5mm}
\subsection{Intermediate Aggregation at HAPS}
For ground clients with weak or lost connectivity to their associated HAPS, local updates are uploaded to nearby BSs. The BS aggregates these updates from all ground clients within its coverage area, reducing the communication load on higher tiers and facilitating more efficient local model convergence. 
Further, if there are no nearby BS available, UAVs can act as relays to transmit local updates to the HAPS. While UAVs are essential for maintaining communication in such scenarios, their limited power supply prevents them from performing any FL-related computation tasks. Instead, they focus solely on relaying data to ensure the continuity of the federated learning process.
\vspace{-3.8mm}
\subsection{Exchange within HAPS Constellation: Distributed FL servers} The HAPS nodes in the HAPS constellation serve as distributed FL servers, responsible for aggregating the cluster-level local updates. Particularly, we assume that HAPS nodes operate independently in a ring topology and exchange their cluster-level local updates with neighbouring nodes. To ensure efficient communication, a strategic approach is adopted, where each HAPS only communicates with its neighbouring nodes within a defined radius. This strategy not only enhances reliability but also mitigates the risk of interference caused by excessive communications within the constellation, maintaining the overall network performance and scalability. Additionally, each HAPS is connected to at least one global relay node, such as a GEO or MEO satellite, ensuring seamless global model exchange with other HAPS constellations across the globe. This connection enables ubiquitous FL implementation within the NTN architecture, facilitating continuous collaboration and model updates on a global scale.
\vspace{-3.8mm}
\subsection{New Global Model Generation, Sharing, and Broadcasting}
After exchanging aggregated cluster-level local updates within the HAPS constellation and receiving global models from different HAPS constellations around the globe via GEO and MEO satellites, each HAPS generates a new global model. This process typically utilizes federated averaging (FedAvg), the most widely used algorithm, or other advanced algorithms tailored to specific learning requirements. Once the new model is generated, each HAPS broadcasts it to two key entities: (i) ground and LEO clients for further cluster-level local training, and (ii) the associated GEO or MEO satellites for global dissemination to other HAPS constellations across the globe.

\vspace{-3.8mm}
\subsection{Leveraging FL Data
for NTN
Management}
Once FL resource allocation and scheduling are optimized and the FL task is complete, i.e., the model converges or maximum communication rounds is reached, the FL data gathered during this process—such as energy usage patterns, resource utilization metrics, and task completion times—can be used to further refine the NTN architecture. This data provides valuable insights into network behaviour, enabling more efficient resource allocation strategies for future communication or specific learning tasks. For example, learning how devices with varying energy profiles contributed to the FL process can inform future scheduling decisions, prioritizing devices with more available energy or resources. Similarly, understanding bandwidth utilization during the FL process can help optimize future bandwidth allocation across LEO, MEO, GEO, and HAPS nodes, ensuring smoother communication and lower latency. This feedback loop of learning and optimization creates a dynamic and adaptive NTN architecture, continuously improving the efficiency and performance of the network based on real-time data generated by the FL process.
\vspace{-3.8mm}
\subsection{FL Communication Model}
For the considered multi-tier NTN, we assume that communication will be performed using FSO links and radio frequency (RF) links. To leverage the strong LoS links between HAPS and satellites, we use the FSO band for communication between satellite and aerial tiers. For intra-aerial tier links, i.e., HAPS-to-HAPS and HAPS-to-UAVs, the FSO band is employed due to strong LoS links. Importantly, there is no interference with HAPS-to-LEO links, as model exchange occurs after the aggregation process.

For communication between aerial and terrestrial tiers, we use RF spectrum bands. Each HAPS communicates with ground devices via RF links for both uploading and broadcasting purposes. Ground clients within the coverage area of a HAPS or a ground BS use the frequency division multiple access (FDMA)  transmission scheme to upload their local updates simultaneously. 

Further, all communications within the terrestrial tier utilize RF links, employing the FDMA transmission scheme, consistent with conventional FL systems.
\vspace{-3.5mm}
\subsection{FL Computation Model}
In our framework, ground and LEO clients generate cluster-level local updates. Ground clients, like smartphones and IoT sensors, vary in computational power and CPU frequency, affecting task efficiency. To conserve energy, they use lightweight algorithms, while LEO clients run low-power training epochs due to energy constraints.

HAPS nodes, serving as FL servers, handle substantial data aggregation and processing with their high computational power and energy resources. They also enable decentralized HFL through communication with neighbouring HAPS nodes, bearing the main computational load.

GEO and MEO satellites, constrained by limited power, do not perform computations. Instead, they relay global models across the HAPS constellations worldwide, ensuring seamless coverage and consistent learning.
\begin{table*}[!t]
\centering
\caption{Simulation Parameters and Considerations}
\vspace{-2mm}
{\rowcolors{2}{white}{black!10}
\begin{tabular}{|c|c|c|c|c|}
\hline
\textbf{Parameter} & \textbf{Terrestrial-only Scenario} & \textbf{Satellite-RF Scenario} & \textbf{Satellite-FSO Scenario} & \textbf{Distributed-HAPS Scenario} \\ \hline
\textbf{Number of FL Clients} & 20 Ground Clients & 5 LEO clients & 5 LEO clients & 200 Ground Clients \& 5 LEO clients \\ \hline
\textbf{FL Server} & 1 BS & 1 HAPS & 1 HAPS & 5 HAPS \\ \hline
\textbf{Channel Quality} & NLoS & LoS & LoS & LoS \\ \hline
\textbf{Communication Type} & RF & RF & FSO & RF and FSO  \\ \hline
\textbf{Communication Rounds} & 10 & 10 & 10 & 10 \\ \hline
\textbf{Channel Delay (ms)} & 5 ms & 50 ms & 10 ms & 8 ms \\ \hline
\textbf{Learning Rate} & 0.01 & 0.01 & 0.01 & 0.01 \\ \hline
\textbf{Batch Size} & 32 & 32 & 32 & 32 \\ \hline
\textbf{Epochs per Round} & 120 & 120 & 120 & 120 \\ \hline
\textbf{FL Model} & FedAvg & FedAvg & FedAvg & FedAvg \\ \hline
\textbf{Loss Function} & Cross-Entropy & Cross-Entropy & Cross-Entropy & Cross-Entropy \\ \hline 
\textbf{Dataset} & non-IID CIFAR-10 & non-IID CIFAR-10 & non-IID CIFAR-10 & non-IID CIFAR-10 \\ \hline
\end{tabular}} \vspace{-2mm}
\label{tab:simulation_parameters}
\end{table*}
\begin{figure*}[!t]
    \hspace*{-3mm}
    \centering 
   \includegraphics[width=175mm]{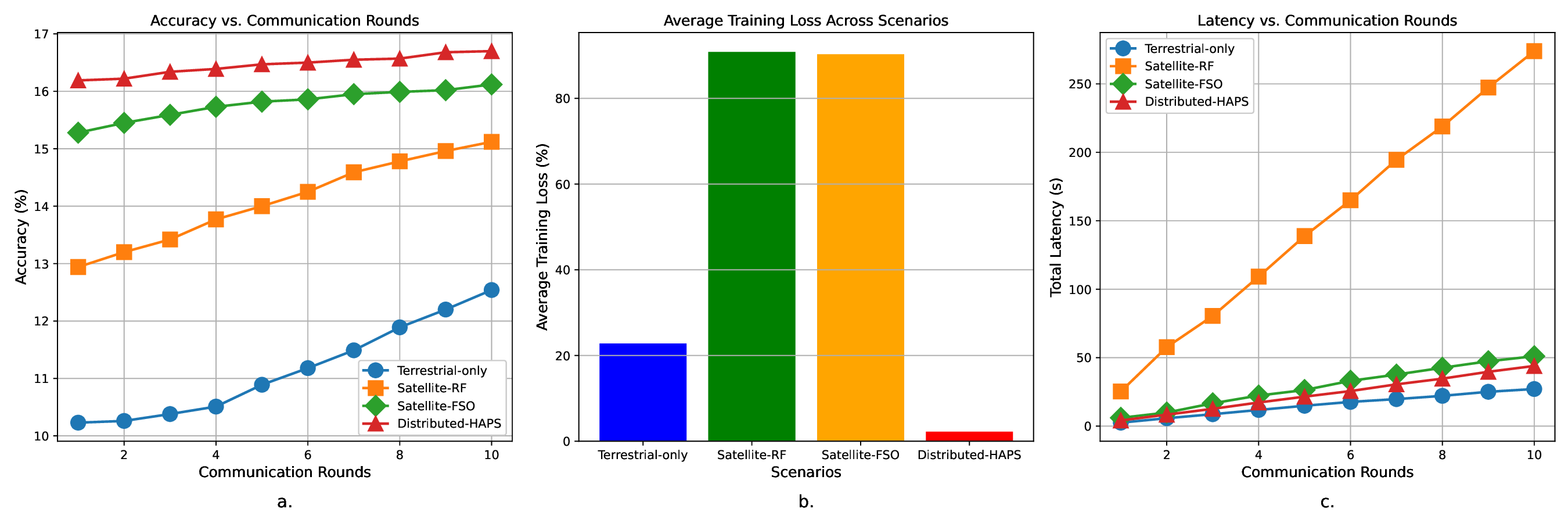}
   \vspace{-4mm}
    \caption{Accuracy, average training loss, and total latency performance comparison of the proposed distributed HFL framework with three baseline frameworks.}
    \vspace{-2.5mm}
    \label{fig3}
\end{figure*}
\vspace{-3mm}
\section{Simulation: A Case Study} In this section, we present preliminary performance results of the proposed distributed HFL framework for NTN in terms of model accuracy, training loss, and latency. Model accuracy measures the performance of the global model, training loss indicates the average loss computed during local training rounds, and latency represents the total communication and computation delays incurred to complete the FL task or meet the predefined stopping criteria. The details of simulation parameters and their values are summarized in TABLE~\ref{tab:simulation_parameters}. We allocated equal transmit power and bandwidth to each client at the physical layer for simplicity. We consider the following four scenarios for evaluation and comparison:
\vspace{-0.6mm}
\begin{itemize} 
\item \textit{Terrestrial-only:} A conventional centralized terrestrial-based FL system is implemented, serving as a baseline for comparison. 
\item \textit{Satellite-RF:} A scenario where LEO satellites, as FL clients, communicate with a single HAPS node using RF communication links. There are no ground clients.
\item \textit{Satellite-FSO:} Similar to the Satellite-RF scenario, but utilizing FSO for communication between satellites and the HAPS node.
\item \textit{Distributed-HAPS:} The proposed framework, with multiple HAPS nodes acting as FL servers, generates new models in a distributed manner by exchanging each HAPS's aggregated cluster-level local updates with all other HAPS nodes.
\end{itemize}

Fig.~\ref{fig3}a compares the accuracy of the four scenarios. The Distributed-HAPS scenario shows significantly higher accuracy due to the larger number of clients, including both ground and satellite nodes. HAPS nodes' strategic positioning and ability to exchange aggregated data enable more efficient model generation and update dissemination, leading to faster convergence and higher accuracy. The Terrestrial-only scenario, with fewer clients, has slower accuracy improvement and plateaus at a lower level. While the Satellite-RF and Satellite-FSO scenarios outperform Terrestrial-only due to broader coverage, they still fall behind Distributed-HAPS. Satellite-RF experiences latency that slows model convergence, and Satellite-FSO, despite lower latency, has fewer participating clients, limiting its performance.

Fig.\ref{fig3}b compares the training loss across the four scenarios. The Distributed-HAPS scenario shows the lowest average training loss, indicating more efficient error minimization and aligning with the higher accuracy observed in Fig.\ref{fig3}a. In contrast, the Terrestrial-only scenario has higher training loss, reflecting challenges in convergence due to fewer clients and reduced data diversity. Both the Satellite-RF and Satellite-FSO scenarios also show higher training loss than Distributed-HAPS, highlighting the advantage of a distributed architecture with broader client participation and efficient communication links.

Fig.~\ref{fig3}c compares the total latency across the four scenarios. The Distributed-HAPS scenario shows higher cumulative latency than the Terrestrial-only scenario, mainly due to the distributed model generation process involving data exchange among multiple HAPS nodes. While this distributed approach adds communication overhead, it improves model accuracy. The Satellite-RF scenario has significantly higher latency compared to Satellite-FSO, highlighting RF communication's limitations in speed and reliability. Although the Distributed-HAPS scenario incurs higher latency, this trade-off is justified by substantial gains in model accuracy and reduced training loss. Overall, the framework delivers high precision and responsiveness, making it ideally suited for applications such as autonomous vehicular networks and remote healthcare monitoring.
\vspace{-3mm}
\section{Open Challenges and Future Directions}
This section highlights key deployment challenges and future directions for integrating distributed HFL into NTN.

\textbf{Limitations of HAPS as FL Server:}
HAPS as an FL server has many benefits, but it also faces some challenges in practical deployments. Limited power and energy can make it challenging to manage complex tasks and scale effectively. Also, environmental factors such as weather conditions and the inherent mobility of aerial platforms may lead to intermittent connectivity and variable performance. Hence, robust resource management strategies must be developed to ensure reliable and efficient FL operations.


\textbf{Efficient Scheduling for FSO Transmissions:} FSO links suffer from interference among multiple concurrent FSO transmissions (e.g., between HAPS and several LEOs or MEO/GEO satellites). Further research is required to create dynamic schedulers that adapt to atmospheric and network conditions to minimize interference and ensure reliable, high-throughput communication for distributed learning.


\textbf{HAPS Constellation Design:} Optimal placement of HAPS nodes is vital for enhancing coverage and operational efficiency within the HFL framework, balancing geographic reach with low latency. Investigating optimal placement, along with developing scalable, adaptive constellation architectures, is essential to meet dynamic network needs and environmental changes.

\textbf{Excessive Communication Overhead:} The multi-tier NTN structure introduces substantial communication overhead, particularly in synchronizing model updates across tiers. Decentralized aggregation methods can minimize the frequency of inter-node communication by enabling each node to perform local aggregation of model updates. Besides, future work should focus on adaptive protocols and compressed update techniques.

\textbf{Coordination and Synchronization:} Coordinating across NTN's heterogeneous tiers is challenging due to varying latencies and node capabilities. Future work should explore advanced hierarchical synchronization protocols and consensus mechanisms—such as Gossip-based algorithms—that dynamically adapt to network conditions, ensuring consistent and reliable model updates across terrestrial, aerial, and satellite tiers.

\textbf{FL-aided Integrated Sensing and Communication (ISAC):} FL-aided ISAC system within NTN helps in reducing the latency by leveraging the dual use of communication signals for both data transmission and sensing. However, this introduces unique challenges including the optimization of resource allocation among sensing, communication, and computation tasks, as well as interference mitigation and balancing sensing accuracy with communication efficiency. Addressing these challenges is critical for enhancing the performance and intelligence of NTN architectures and will pave the way for more robust and efficient future networks.
\vspace{-2mm}
\section{Conclusion}
With its multi-tier structure—encompassing terrestrial, aerial, and satellite tiers—NTN will play a pivotal role in 6G and beyond. This article explores the suitability of FL for NTNs and how NTNs can enhance FL performance. We reviewed FL in NTNs and proposed the HAPS constellation as the ideal choice for distributed FL mechanisms. Our proposed distributed HFL framework improves model accuracy and reduces training loss, with simulation results confirming its superiority, despite a trade-off in latency. These preliminary findings highlight the importance of distributed architectures for extensive client participation and efficient communication in large-scale, heterogeneous NTNs. The proposed framework shows great potential for enhancing FL performance in future NTN architectures.
\vspace{-2mm}
\bibliographystyle{IEEEtran}

\begin{IEEEbiographynophoto}{Amin Farajzadeh}
\textbf{[SM]} (aminfarajzadeh@sce.carleton.ca) is currently a Ph.D., candidate in the Carleton Non-Terrestrial
Networks Lab in the Systems and Computer Engineering Department at Carleton University, Canada. 
\end{IEEEbiographynophoto}
\vspace{-1mm}
\vskip -2\baselineskip plus -1fil
\begin{IEEEbiographynophoto}{Animesh Yadav}
\textbf{[SM]} (yadava@ohio.edu) is an assistant professor in the School of Electrical Engineering and Computer Science at Ohio University, USA. 
\end{IEEEbiographynophoto}
\vspace{-1mm}
\vskip -2\baselineskip plus -1fil
\begin{IEEEbiographynophoto}{Halim Yanikomeroglu}
\textbf{[F]} (halim@sce.carleton.ca) is a chancellor’s professor in
the Department of Systems and Computer Engineering at Carleton University, Canada, and director of the Carleton Non-Terrestrial Networks
Lab. 
\end{IEEEbiographynophoto}
\end{document}